\documentclass[a4paper,twoside,12pt]{article}

\usepackage{multicol} 
\usepackage{graphicx}
\usepackage{amsmath}
\usepackage{mathptmx}
\usepackage{hyperref}
\usepackage{amsthm}
\usepackage{amsfonts}
\usepackage{amssymb}
\usepackage{bm}
\usepackage[english]{babel}
\usepackage[utf8]{inputenc}
\usepackage{color,colortbl,multirow}
\definecolor{Gray}{gray}{0.8}
\usepackage{tabularx}
\usepackage[table]{xcolor}
\definecolor{Gray}{gray}{0.9} 
\newcolumntype{W}[1]{>{\hsize=#1\hsize\raggedleft\arraybackslash}X} 
\newcolumntype{C}[1]{>{\hsize=#1\hsize\centering\arraybackslash}X}  
\usepackage{caption}
\usepackage{array}
\usepackage{subfigure}
\usepackage{placeins}
\usepackage{float}
\usepackage{dcolumn}

\usepackage[T1]{fontenc}

\usepackage[hmarginratio=1:1,top=32mm,columnsep=20pt]{geometry} 
\usepackage{abstract} 
\usepackage{fancyhdr}
\title{Classification of Transient Astronomical Object Light Curves Using LSTM Neural Networks}

\author{
Fernandes, G.G.D.$^{1,3}$; Barroca, M.A.$^1$; dos Santos, M.$^1$; Oliveira, R.S.$^2$\\ 
\footnotesize $^1$Centro Brasileiro de Pesquisas Físicas \\
\footnotesize $^2$Universidade Federal de Minas Gerais \\
\footnotesize $^3$Instituto Superior Técnico \\ 
\vspace{-0mm}
}

\date{~~~}
\begin{document}

\maketitle 

\thispagestyle{fancy} 

\vspace{-2.0cm}

\begin{abstract}

\vspace{-3mm}
\noindent This study presents a bidirectional Long Short-Term Memory (LSTM) neural network for classifying transient astronomical object light curves from the Photometric LSST Astronomical Time-series Classification Challenge (PLAsTiCC) dataset. The original fourteen object classes were reorganized into five generalized categories (S-Like, Fast, Long, Periodic, and Non-Periodic) to address class imbalance. After preprocessing with padding, temporal rescaling, and flux normalization, a bidirectional LSTM network with masking layers was trained and evaluated on a test set of 19,920 objects. The model achieved strong performance for S-Like and Periodic classes, with ROC area under the curve (AUC) values of 0.95 and 0.99, and Precision-Recall AUC values of 0.98 and 0.89, respectively. However, performance was significantly lower for Fast and Long classes (ROC AUC of 0.68 for Long class), and the model exhibited difficulty distinguishing between Periodic and Non-Periodic objects. Evaluation on partial light curve data (5, 10, and 20 days from detection) revealed substantial performance degradation, with increased misclassification toward the S-Like class. These findings indicate that class imbalance and limited temporal information are primary limitations, suggesting that class balancing strategies and preprocessing techniques focusing on detection moments could improve performance.
\end{abstract}

\vspace{8mm}
\begin{multicols}{2} 
\section{Introduction}
Artificial intelligence (AI) has become increasingly important in scientific research, particularly for analyzing large-scale datasets that exceed human processing capabilities. In astronomy, the exponential growth of observational data from modern telescopes requires automated classification systems to identify and categorize celestial objects efficiently.\cite{Djorgovski_2022} Machine learning approaches, particularly deep neural networks, have shown great promise in handling such classification tasks.

Recently, the Large Synoptic Survey Telescope (LSST) proposed a challenge on Kaggle, a Google subsidiary website, called The Photometric LSST Astronomical Time-series Classification Challenge (PLAsTiCC).\cite{kaggle} This challenge is aimed at classifying light curves, data of the observed brightness of celestial objects as a function of time, which were simulated in preparation for LSST observations. These curves are capable of revealing the presence of some phenomena, such as supernovae\cite{Russel_1912}. The LSST will revolutionize our understanding of the sky, however, this type of study is hindered by the large volume of data obtained by the telescope, making automatic analysis procedures indispensable to differentiate and classify them. In this challenge, the question was raised: How well can we classify objects in the sky that vary in brightness from simulated LSST time-series data? The PLAsTiCC dataset and challenge were created to help classify these astronomical sources. \cite{PLAsTiCC}

Before the use of AI, this type of classification still depended on manual analysis. Common statistical methods such as template fitting\cite{Tanvir_2005} were used, but these are not scalable for a large volume of data. Usually, a group of experts manually eliminated obvious cases of false positives, something that by itself can take several days. Of the remaining data, each case should be reviewed by at least three experts, which can lead to disagreements about a particular case, since experts might not have the same definition for classification. For these reasons, a reliable system is necessary that repeatedly selects the most important candidates that will be manually reviewed for confirmation at a later stage.

\section{Theoretical Framework}

\subsection{Long Short-Term Memory Neural Networks}

\begin{figure*}[ht]
    \centering
    \includegraphics[width=0.75\textwidth]{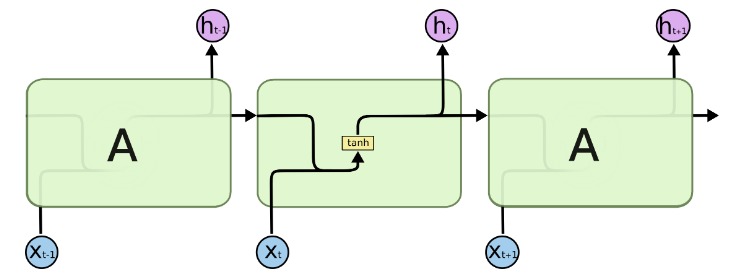}
    \caption{Functioning of an RNN network. \cite{colah}}
    \label{fig:RNN}
\end{figure*}

\begin{figure*}[ht]
    \centering
    \includegraphics[width=0.75\textwidth]{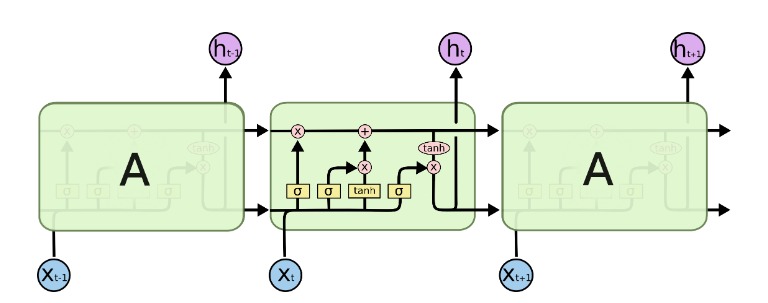}
    \caption{Functioning of an LSTM network. \cite{colah}}
    \label{fig:lstm}
\end{figure*}

Since light curves are a function that varies with time, one possible way to solve the PLAsTiCC problem is to use a Long Short-Term Memory (LSTM) Neural Network. The LSTM is a recurrent neural network (RNN) architecture, developed by Hochreiter and Schmidhuber in 1997. Recurrent networks have the characteristic of making some information persist in the network through a loop at arbitrary intervals. The RNN by definition has a simpler repetition structure than LSTMs, as there is only a single tanh layer, as shown in figure \ref{fig:RNN}, while LSTMs have a chain structure that contains four neural network layers and different memory blocks called cells as shown in figure \ref{fig:lstm}. Thus, LSTMs are well suited for classifying, processing, and predicting time series with time intervals of unknown duration. \cite{Gabriel_2022}

\section{Materials and Methods}

The light curve data were obtained through LSST PLAsTiCC via the Kaggle challenge\cite{kaggle}. To complete the project within the scheduled time of EAFEXP, a subset of the total data was used. In total, two datasets were used: the first had 7848 objects, of which 10$\%$ were used to form a validation set and the remainder formed the training set; the second dataset had 19920 objects and was used as the test set.

The data are in tabular format and consist of flux measurement information, measurement error, the measurement time in Modified Julian Date (mjd), the filter used (the object's emission spectrum, ugriZY, represented by a numerical value from zero to five), a boolean variable (0 or 1) that indicates detection or non-detection, the object class, and a unique identifier. An example can be seen in Table \ref{tab:conjunto-dados-ex}.

\begin{center}
\begin{minipage}{\columnwidth}
\centering
\setlength{\tabcolsep}{2pt}     
\renewcommand{\arraystretch}{1.1}
\footnotesize                   
\resizebox{\textwidth}{!}{
\begin{tabular}{|r|r|r|c|c|c|r|}
\hline
\textbf{Flux} & \textbf{Error} & \textbf{mjd} & \textbf{Filter} & \textbf{Detection} & \textbf{Class} & \textbf{Id} \\\hline
-544.810 & 3.623  & 59750.4 & 2 & 1 & 92 & 615 \\\hline
-816.434 & 5.553  & 59750.4 & 1 & 1 & 88 & 713 \\\hline
-471.386 & 3.802  & 59750.4 & 3 & 1 & 42 & 730 \\\hline
... & ... & ... & ... & ... & ... & ... \\\hline
-388.985 & 11.395 & 59750.4 & 4 & 1 & 90 & 745 \\\hline
-2.940   & 1.771  & 59798.3 & 3 & 0 & 90 & 116720 \\\hline
-12.810  & 5.380  & 59798.4 & 5 & 0 & 92 & 117016 \\\hline
... & ... & ... & ... & ... & ... & ... \\\hline
\end{tabular}}
\captionof{table}{Table exemplifying the dataset.}
\label{tab:conjunto-dados-ex}
\end{minipage}
\end{center}

\subsection{Preprocessing}

\begin{figure*}[ht]
    \centering
    \subfigure[Distribution of measurements by original classes.]{\includegraphics[width=0.49\textwidth]{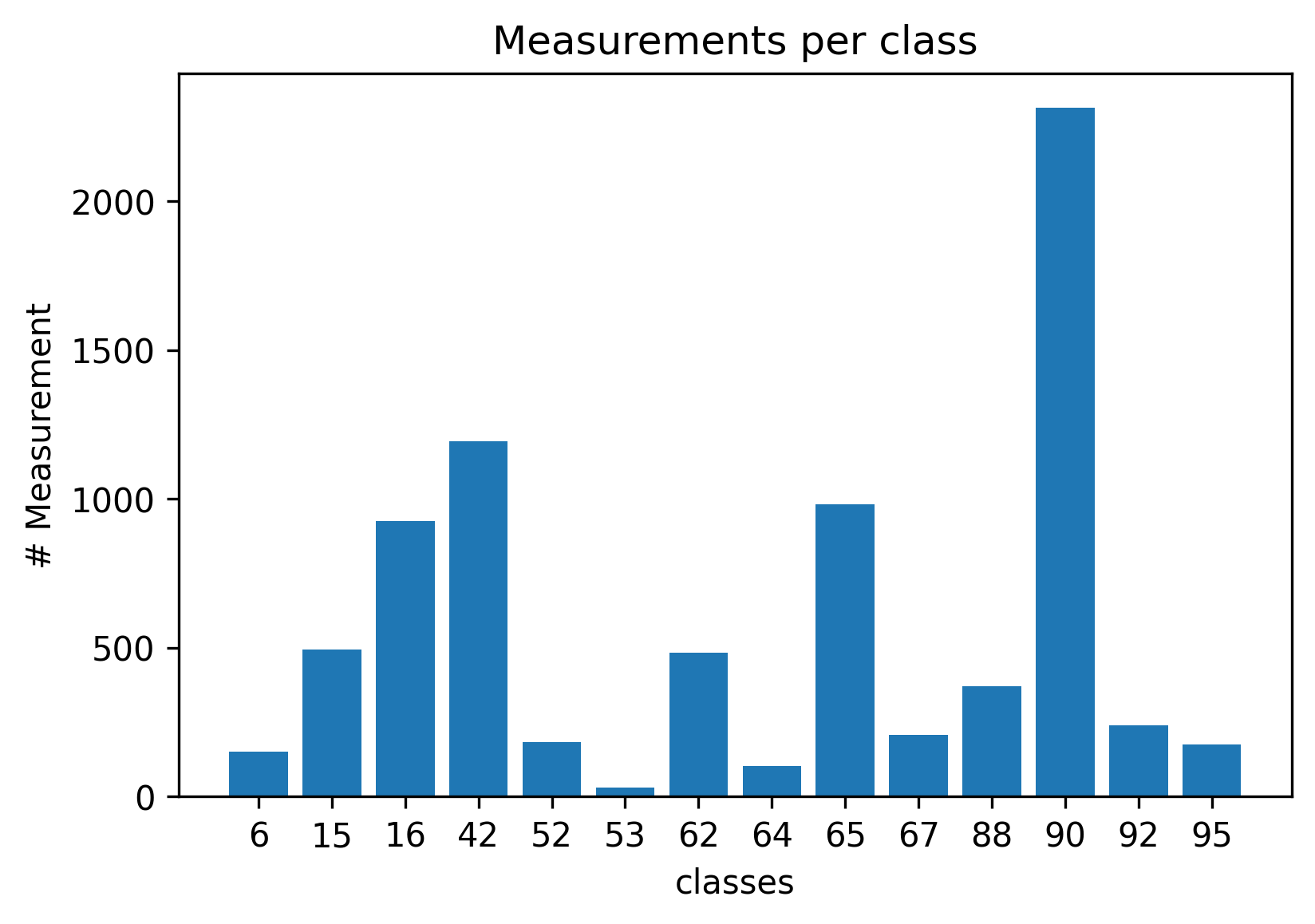}}
    \hfill
    \subfigure[Distribution of measurements by generalized classes.]{\includegraphics[width=0.49\textwidth]{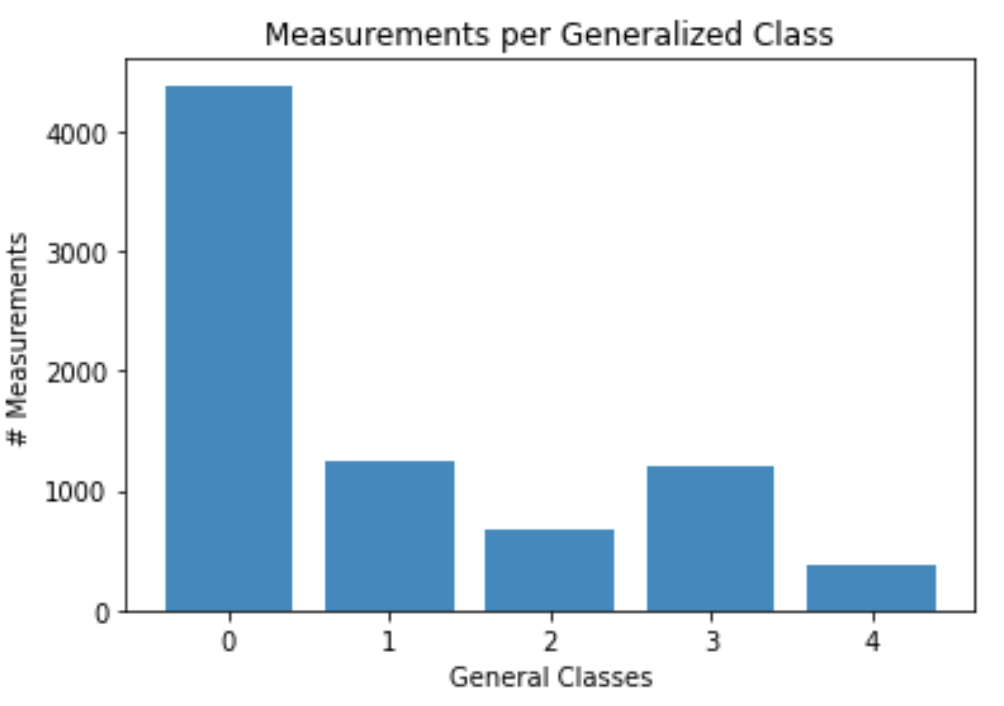}}
    \caption{}
    \label{fig:classes}
\end{figure*}

\begin{figure*}[ht]
    \centering
    \subfigure[Example of a light curve of an object from the S-Like class. The values in the legend indicate the filter used and the emission spectrum.]{\includegraphics[width=0.49\textwidth]{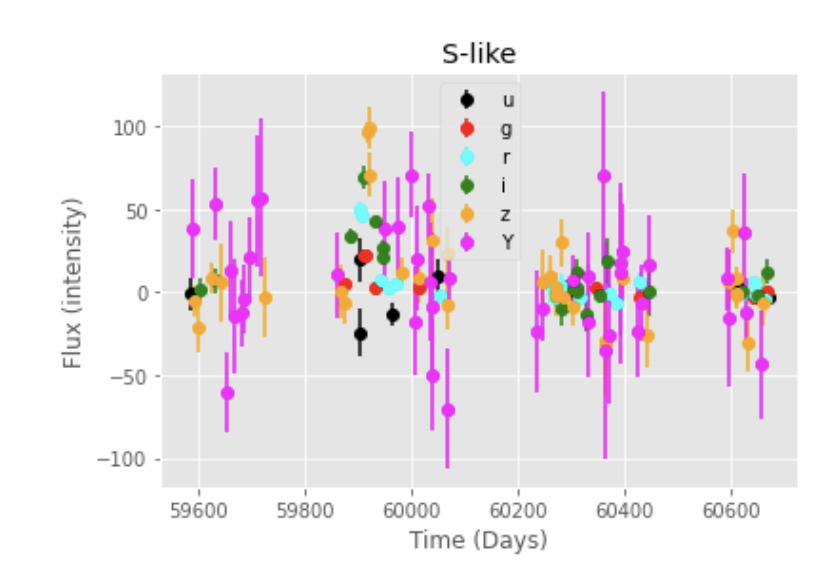}}
    \hfill
    \subfigure[Example of a light curve of an object from the Fast class. The values in the legend indicate the filter used and the emission spectrum.]{\includegraphics[width=0.49\textwidth]{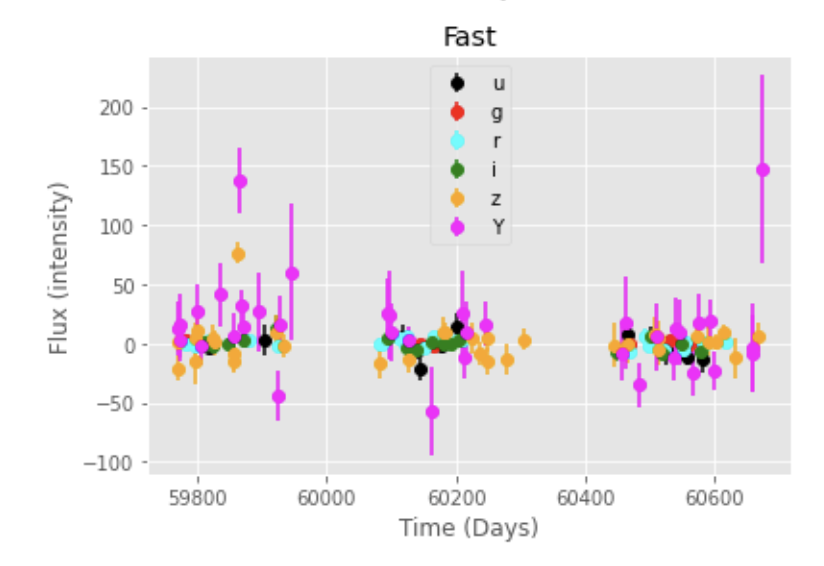}}
    \hfill
    \subfigure[Example of a light curve of an object from the Long class. The values in the legend indicate the filter used and the emission spectrum.]{\includegraphics[width=0.49\textwidth]{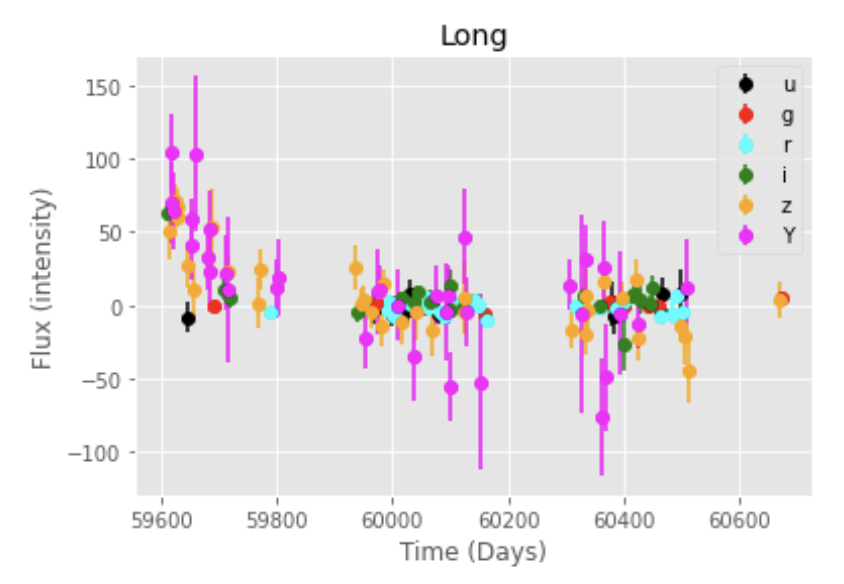}}
    \hfill
    \subfigure[Example of a light curve of an object from the Periodic class. The values in the legend indicate the filter used and the emission spectrum.]{\includegraphics[width=0.49\textwidth]{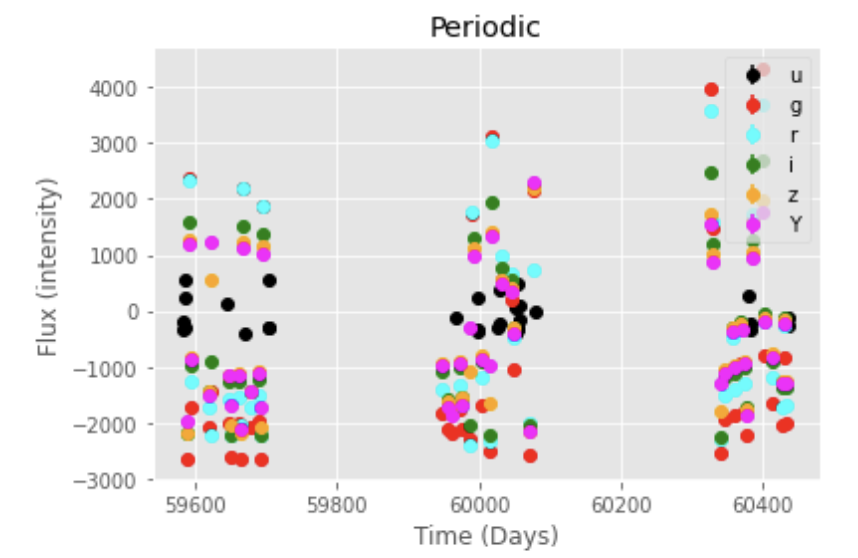}}
    \hfill
    \subfigure[Example of a light curve of an object from the Non-Periodic class. The values in the legend indicate the filter used and the emission spectrum.]{\includegraphics[width=0.49\textwidth]{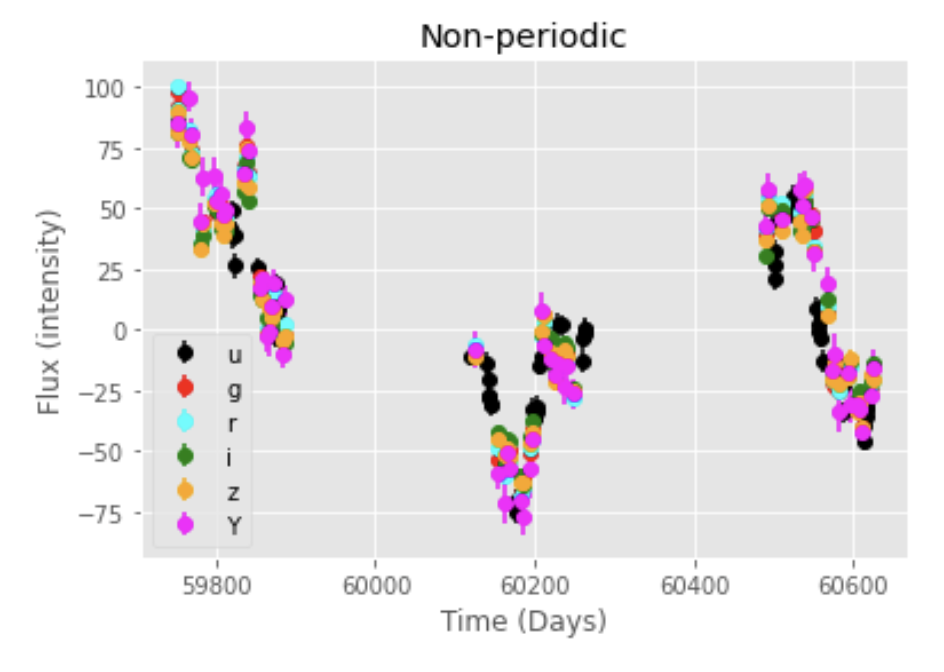}}
    \caption{}
    \label{fig:curvas}
\end{figure*}

Initially, the provided data had fourteen categories, however, due to the discrepancy in the amount of data per category, the data were reorganized into five categories according to Table \ref{tab:classes-generalizadas}. Examples of each of the classes can be seen in figure \ref{fig:curvas}. The distribution by original classes and generalized classes can be seen in figure \ref{fig:classes}.

\begin{center}
\begin{minipage}{\columnwidth}
\centering
\setlength{\tabcolsep}{3pt}
\renewcommand{\arraystretch}{1.1}
\footnotesize
\resizebox{\textwidth}{!}{%
\begin{tabular}{|r|l|l|r|}
\hline
\rowcolor{gray!20}\textbf{Id} & \textbf{Original Class} & \textbf{Generalized Class} & \textbf{New Id} \\\hline
6  & Single micro-lens & Fast         & 1 \\\hline
15 & TDE               & Long         & 2 \\\hline
16 & Eclipsing Binary  & Periodic     & 3 \\\hline
42 & SNII              & S-Like       & 0 \\\hline
52 & SNIax             & S-Like       & 0 \\\hline
53 & Mira              & Periodic     & 3 \\\hline
62 & SNIbc             & S-Like       & 0 \\\hline
64 & Kilonova          & Fast         & 1 \\\hline
65 & M-dwarf           & Fast         & 1 \\\hline
67 & SNIa-91bg         & S-Like       & 0 \\\hline
88 & AGN               & Non-Periodic & 4 \\\hline
90 & SNIa              & S-Like       & 0 \\\hline
92 & RR Lyrae          & Periodic     & 3 \\\hline
95 & SLSN-I            & Long         & 2 \\\hline
\end{tabular}}
\captionof{table}{Table with the fourteen original classes and their generalized classes.}
\label{tab:classes-generalizadas}
\end{minipage}
\end{center}

\begin{figure*}[ht]
    \centering
    \subfigure[Distribution of the number of measurements per object.]{\includegraphics[width=0.49\textwidth]{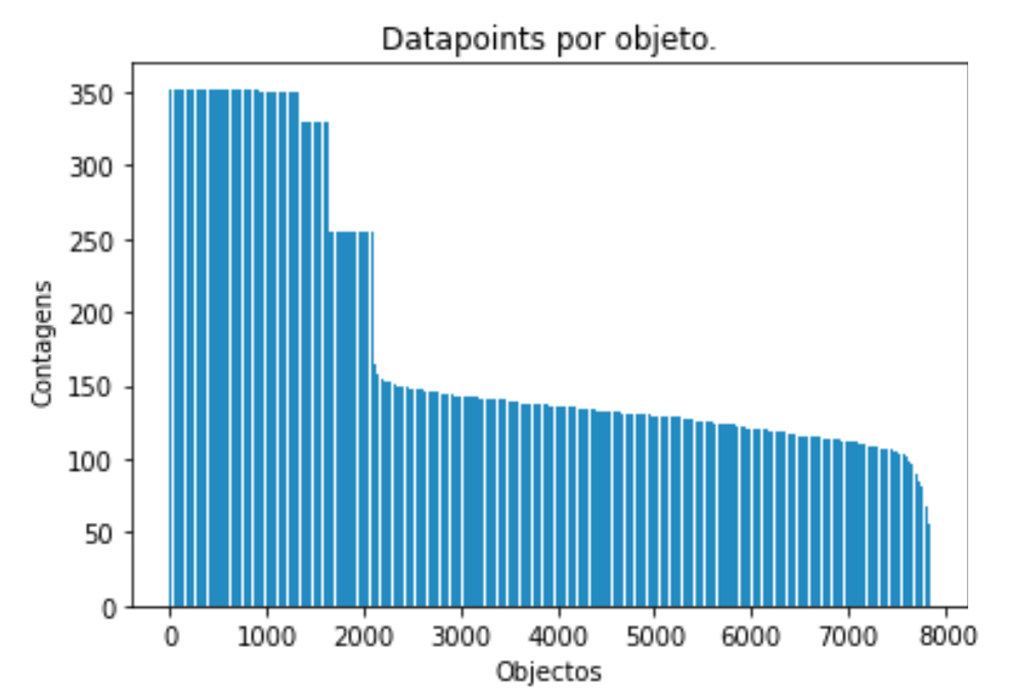}}
    \hfill
    \subfigure[Distribution of the number of measurements per object after padding.]{\includegraphics[width=0.49\textwidth]{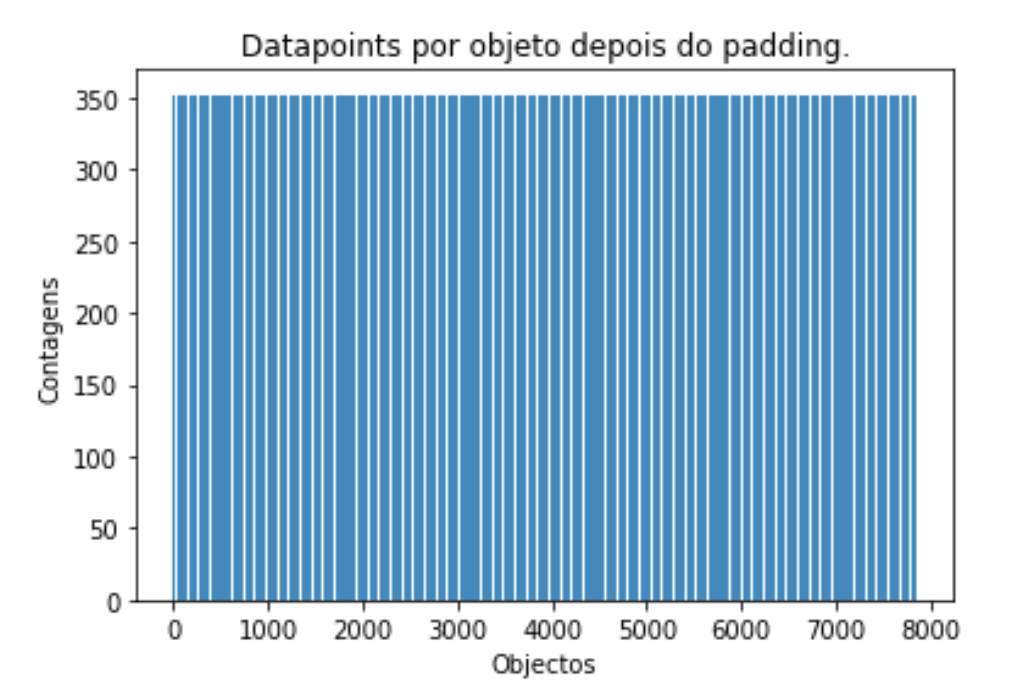}}
    \caption{}
    \label{fig:padding}
\end{figure*}

To train the network, it is necessary to ensure that each object has the same number of measurements, since this will be the input dimension. This is not true for this dataset, so the data with smaller dimensions were filled with arbitrary values in a process known as padding. As shown in figure \ref{fig:padding}, after padding, all objects have the same dimension.

In addition, the temporal measurements were rescaled so that the first measurement of each object was set to zero on the temporal scale. Finally, the flux data were normalized for each object separately using min-max normalization, scaling values to the range [0,1], since the amplitude of measurements varies significantly between objects.

Finally, this project is interested in verifying the ability to classify objects with only partial light curve data. For this purpose, two new test sets were selected, where the data were advanced in time by ten and twenty days of observation from the moment each object is detected, and subsequently the network was evaluated with sets of five, ten, and twenty days into the future.

\subsection{Neural Network}

Since the provided data are time series, an LSTM network was chosen for its ability to capture temporal dependencies. The network is bidirectional to capture both forward and backward temporal patterns, as both previous and subsequent measurements are relevant for classification. Each column in the dataset (Table \ref{tab:conjunto-dados-ex}) was treated as a feature, resulting in five input features: flux, error, modified Julian date, filter, and detection.

Due to the padding procedure, all objects have 352 measurements and 5 features. The network input dimensions are therefore 352$\times$5, and the output is a probability distribution over the five generalized classes.

\begin{figure*}[ht]
    \centering
    \includegraphics[width=0.75\textwidth]{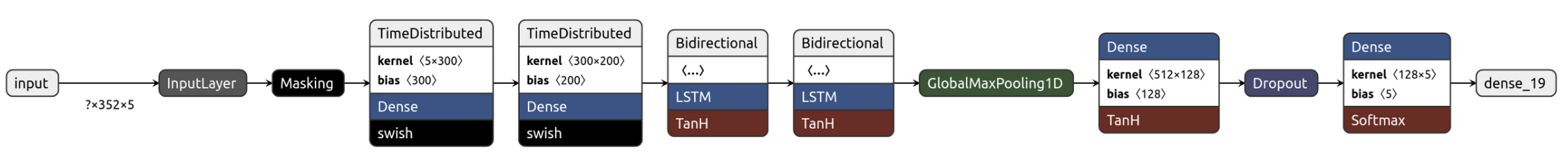}
    \caption{Neural network used in the classification process. The LSTM network is bidirectional because equal importance must be given to past and future data; the Mask layer serves to attenuate the noise effects that padding can cause}
    \label{fig:rede}
\end{figure*}

A masking layer was implemented to identify and ignore padded values, preventing them from being treated as noise. A GlobalMaxPooling layer was used to reduce the temporal dimensions by extracting the maximum value across the sequence, which helps capture the most significant features regardless of their temporal position. Finally, a dense layer with softmax activation performs the final classification. The complete network architecture can be visualized in figure \ref{fig:rede}.

The network was trained using the Adam optimizer with categorical cross-entropy as the loss function. Training was performed with early stopping based on validation loss to prevent overfitting. The model was trained for a maximum of 50 epochs with a batch size of 32.

All code used and the trained model to reproduce these results are available in a public repository on GitHub at the link \url{https://github.com/MarcoBarroca/VI_EAFEXP_Proj3}.

\section{Results and Discussion}

\subsection{Performance on Complete Light Curves}

The network was evaluated using ROC curves, Precision-Recall curves, and confusion matrices, all constructed from the test set. The ROC curves showed excellent performance for the S-Like and Periodic classes, achieving area under the curve (AUC) values of 0.95 and 0.99, respectively. However, the Long class achieved a significantly lower AUC of 0.68, indicating poor performance for this category.

\begin{figure*}[ht]
    \centering
    \subfigure[ROC curve for our classification model. The Periodic and S-Like classes show satisfactory results, however the same cannot be said for the Fast and Long classes.]{\includegraphics[width=0.49\textwidth]{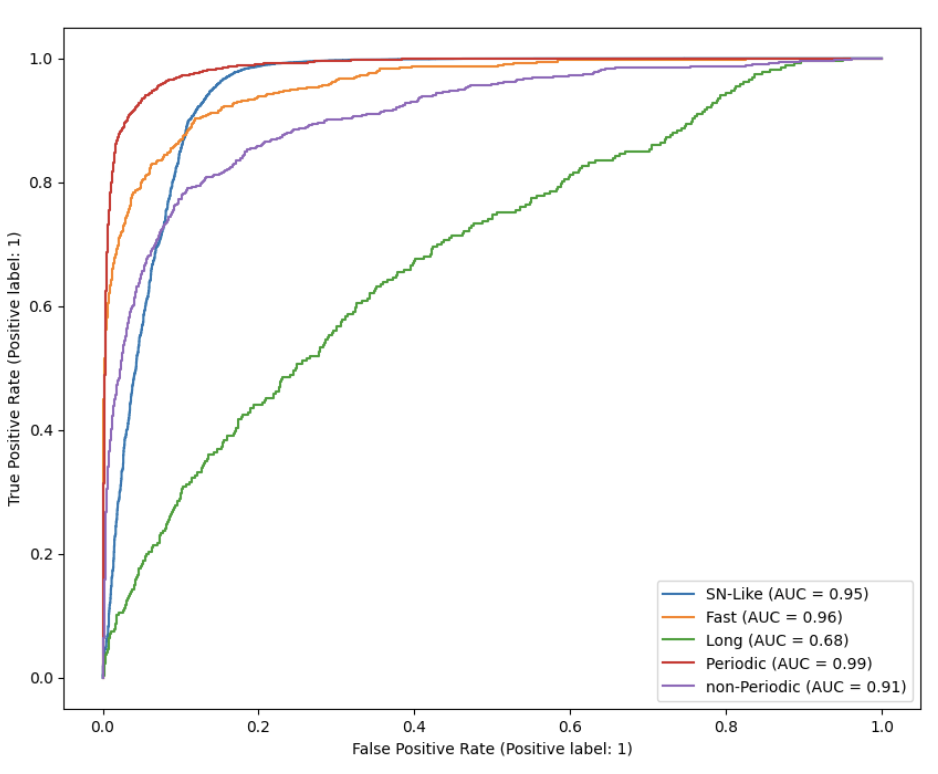}}
    \hfill
    \subfigure[Precision-Recall curve for our classification model. The Periodic and S-Like classes show satisfactory results, however we notice that the model does not seem to correctly classify Non-Periodic curves. ]{\includegraphics[width=0.49\textwidth]{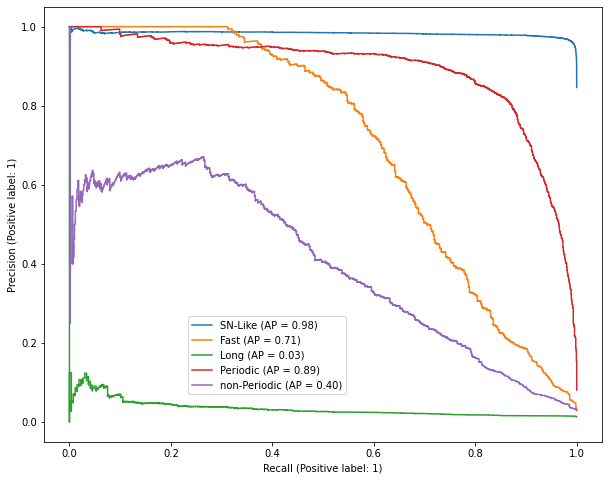}}
    \caption{}
    \label{fig:RoC_Recall}
\end{figure*}

Similarly, the Precision-Recall curves for the S-Like and Periodic classes achieved high AUC values of 0.98 and 0.89, respectively. The Non-Periodic class showed the worst performance with an AUC of 0.40. All curves can be visualized in figure \ref{fig:RoC_Recall}.

Analysis of the confusion matrix (figure \ref{fig:matrizes} (a)) reveals several important patterns. The classification of S-Like curves achieves high accuracy, and the model successfully distinguishes Periodic and Non-Periodic classes from the other three categories. However, the model struggles significantly with Long and Fast type curves, showing poor classification performance. Additionally, the model has difficulty differentiating between Periodic and Non-Periodic objects, often confusing one for the other.

\subsection{Performance on Partial Light Curves}

\begin{figure*}[ht]
    \centering
    \subfigure[Confusion matrix for the complete test set.]{\includegraphics[width=0.49\textwidth]{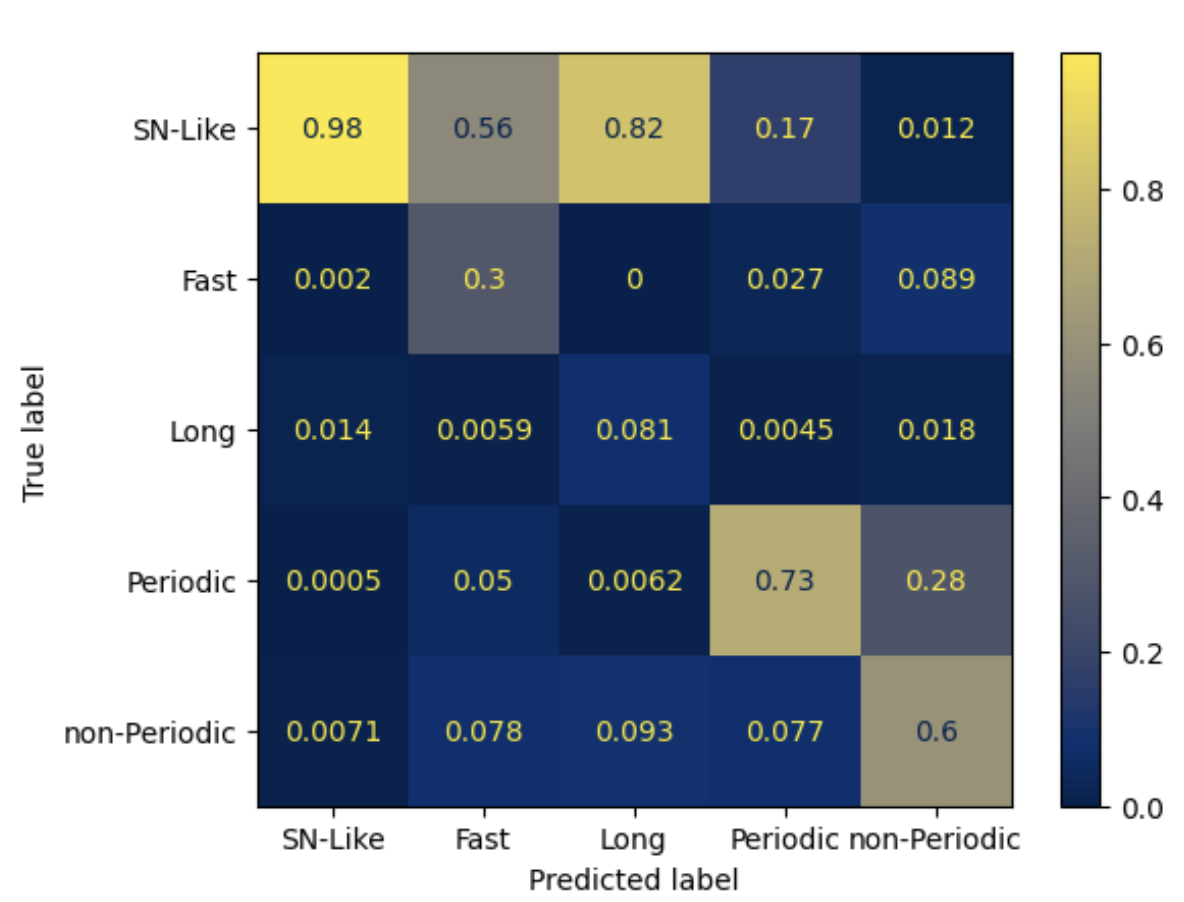}}
    \hfill
    \subfigure[Confusion matrix for the test set with five days ahead. The results worsen compared to the complete test set.]{\includegraphics[width=0.49\textwidth]{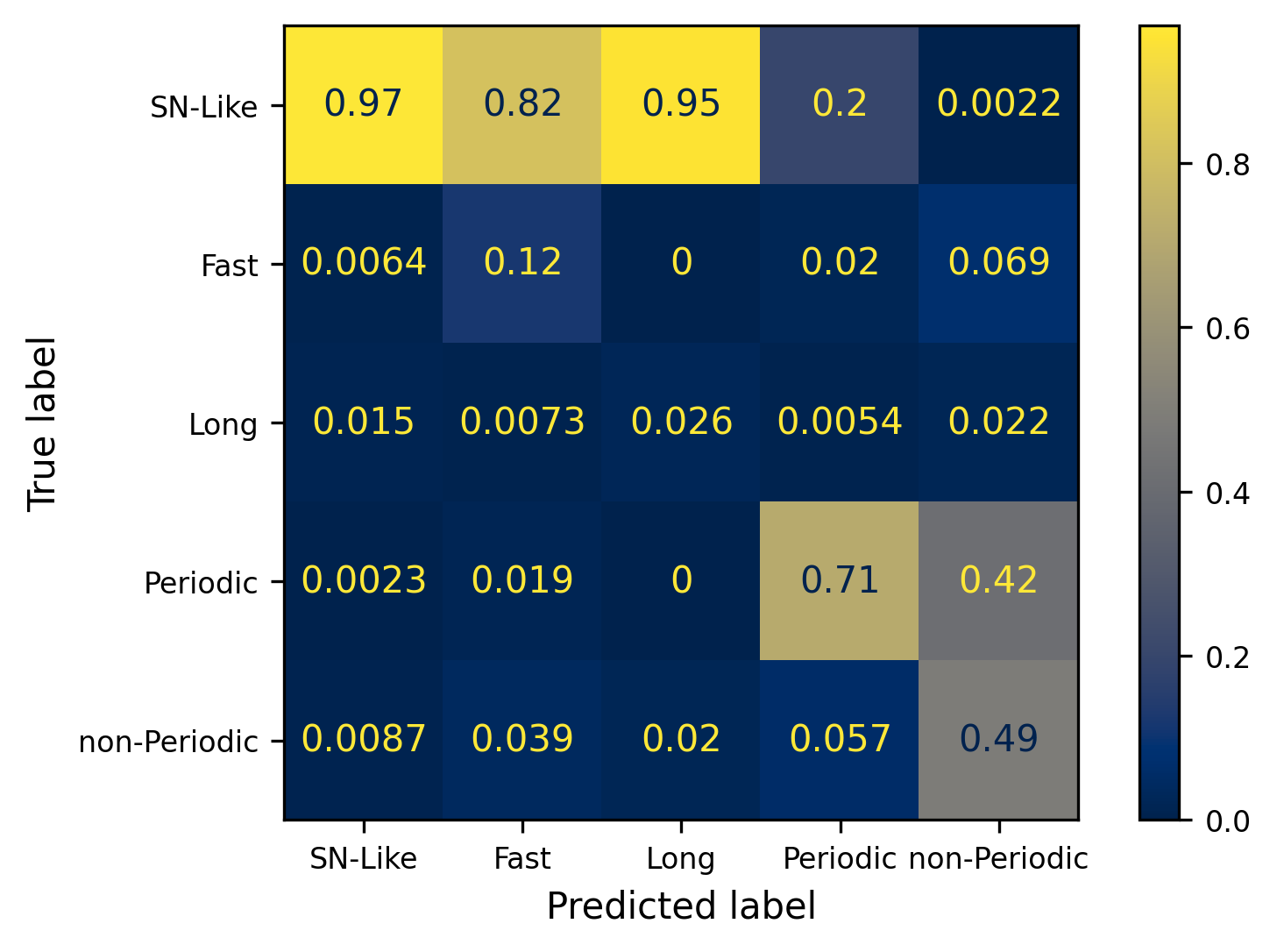}}
    \hfill
    \subfigure[Confusion matrix for the test set with data ten days ahead. The results worsen compared to the complete set and the set with five days.]{\includegraphics[width=0.49\textwidth]{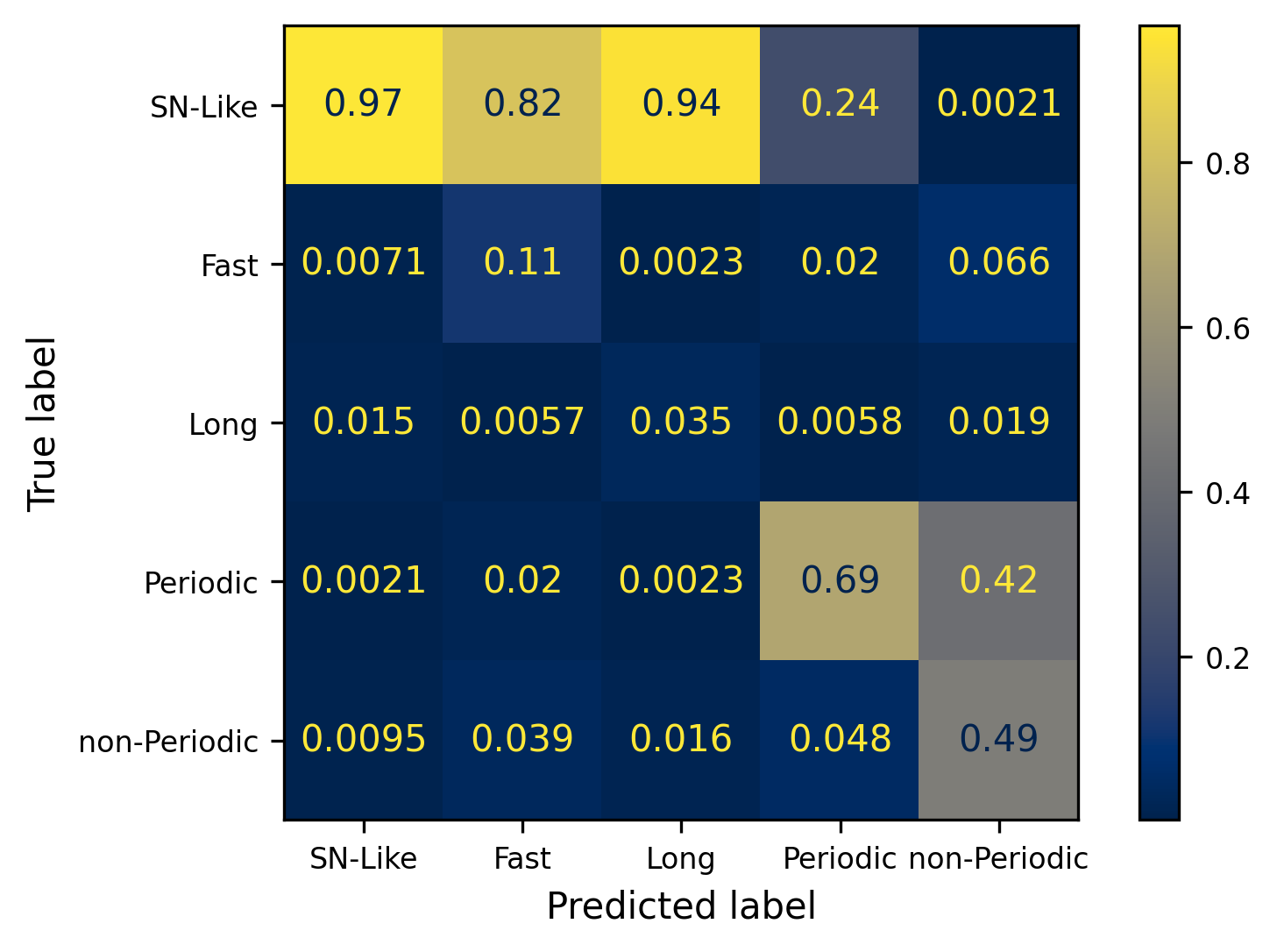}}
    \hfill
    \subfigure[Confusion matrix for the test set with data twenty days ahead. The network now confuses more classes with S-Like type objects.]{\includegraphics[width=0.49\textwidth]{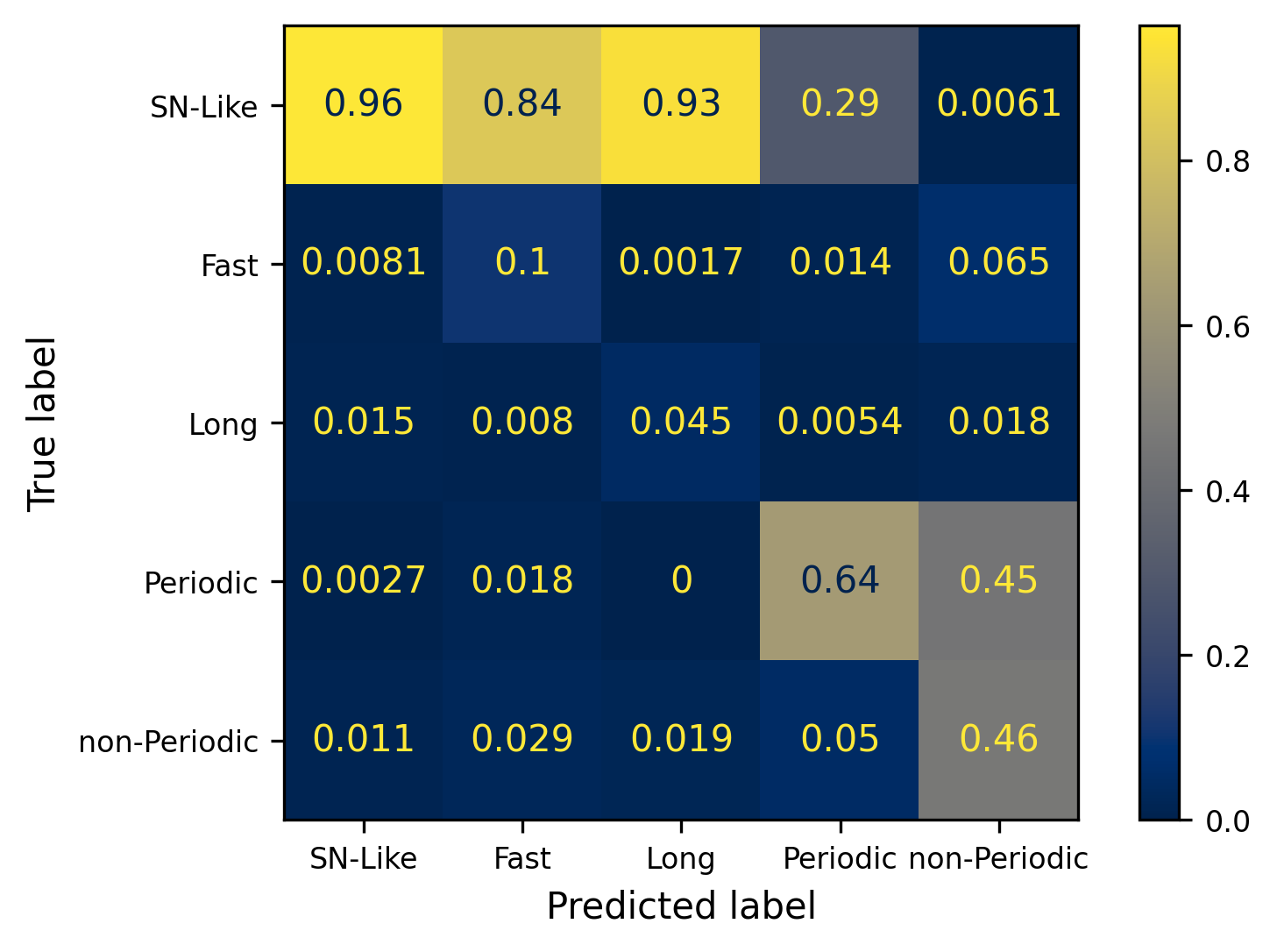}}
    \caption{}
    \label{fig:matrizes}
\end{figure*}

To assess the model's ability to classify objects early in their light curve evolution, the network was evaluated on test sets with temporal advances of 5, 10, and 20 days from the detection moment. The results show a significant degradation in performance as the temporal advance increases. Even with only 5 days of advance, the network fails to achieve performance comparable to the complete dataset and begins to misclassify most objects as S-Like. This degradation becomes more pronounced with longer temporal advances, as shown in the confusion matrices in figure \ref{fig:matrizes} (b), (c), and (d). The ROC and Precision-Recall curves for these partial datasets are provided in the Appendix (figures \ref{fig:roc e precision_5}, \ref{fig:roc e precision_10}, and \ref{fig:roc e precision_20}).

\section{Conclusion}

This work presents a bidirectional LSTM neural network for classifying transient astronomical object light curves from the PLAsTiCC dataset. The model achieved strong performance for S-Like and Periodic classes, with ROC AUC values of 0.95 and 0.99, respectively, demonstrating the effectiveness of LSTM networks for time-series classification in astronomy.

However, several limitations were identified. The model struggled significantly with Fast and Long classes, likely due to their characteristic short-duration peaks that occur only at the beginning or end of measurements, making them difficult to capture with the current approach. Additionally, the model had difficulty distinguishing between Periodic and Non-Periodic objects, often misclassifying Non-Periodic as Periodic, though the reverse error was rare. This suggests that both classes exhibit similar peak patterns, leading the model to incorrectly attribute periodicity to non-periodic signals.

The model's performance degraded substantially when evaluated on partial light curve data, with even 5-day advances causing significant misclassification, primarily toward the S-Like class. This degradation is likely due to the class imbalance in the dataset, where S-Like objects dominate the training distribution.

Several approaches could address these limitations. First, class balancing through oversampling underrepresented classes or using more sophisticated class weighting schemes could improve performance on minority classes. Second, preprocessing strategies that focus on detection moments (using only data where the detection variable equals 1) could help capture the characteristic features of Fast and Long objects while reducing background noise. Third, alternative architectures such as attention mechanisms or transformer networks might better capture long-range dependencies and distinguish periodic from non-periodic patterns. During this work, class weighting was attempted but did not yield significant improvements, suggesting that more substantial architectural or preprocessing changes may be necessary.

\section{Acknowledgements} 

We would like to thank the EAFEXP coordination, professors Clécio R. de Bom, Elisangela L. Faria, Ana Paula O. Muller, and Marcelo Portes de Albuquerque, and the teaching assistants Gabriel Teixeira and Bernardo M. Fraga for the lectures, classes, and assistance with the project.

\end{multicols}

\appendix
\section{Appendix}

\begin{figure}[!h]
    \centering
    \subfigure[Precision-Recall curve for the dataset with a temporal advance of 5 days.]{\includegraphics[width=0.5\textwidth]{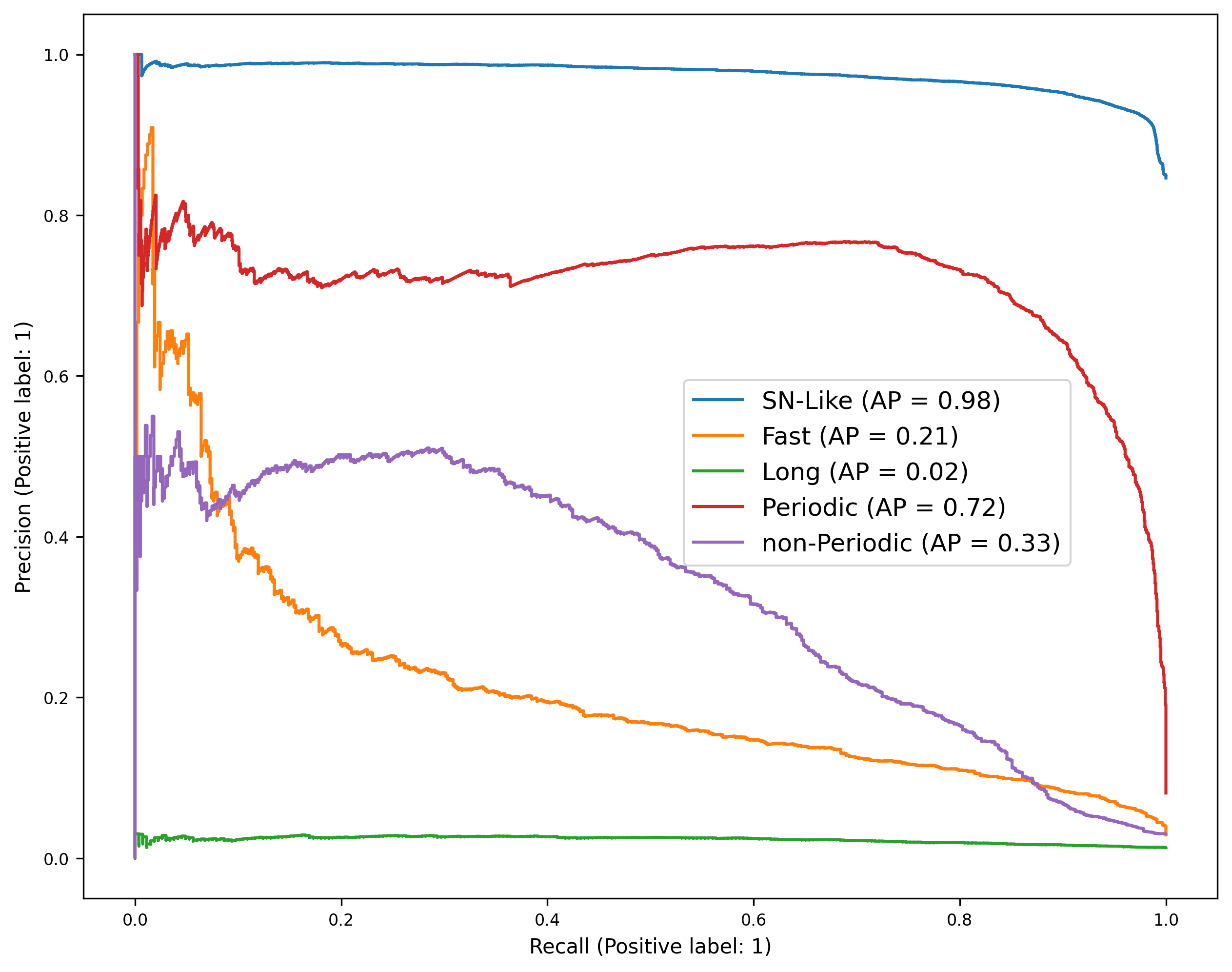}}
    \hfill
    \subfigure[ROC curve for the dataset with a temporal advance of 5 days.]{\includegraphics[width=0.5\textwidth]{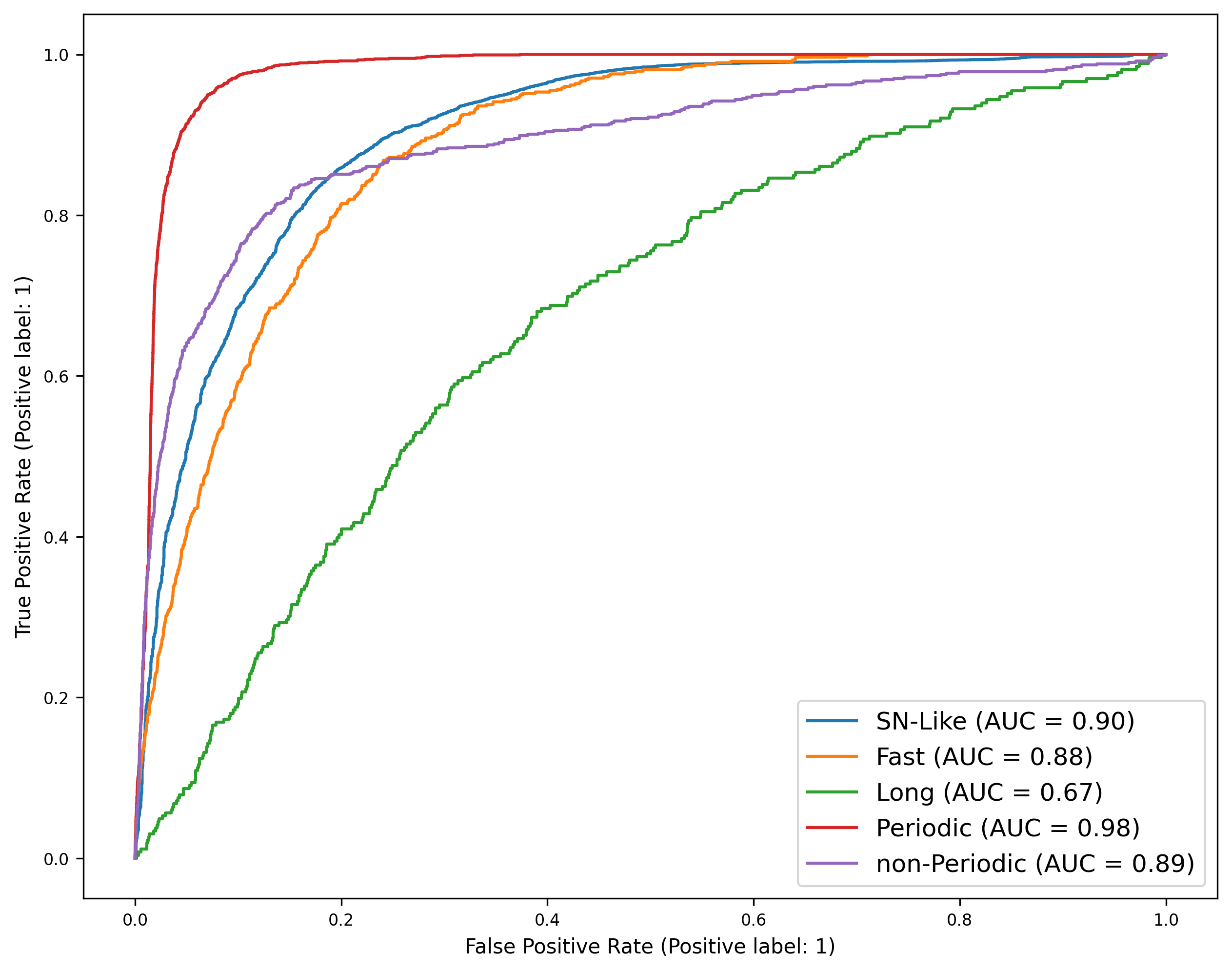}}
    \caption{}
    \label{fig:roc e precision_5}
\end{figure}

\begin{figure}[!h]
    \centering
    \subfigure[Precision-Recall curve for the dataset with a temporal advance of 10 days.]{\includegraphics[width=0.5\textwidth]{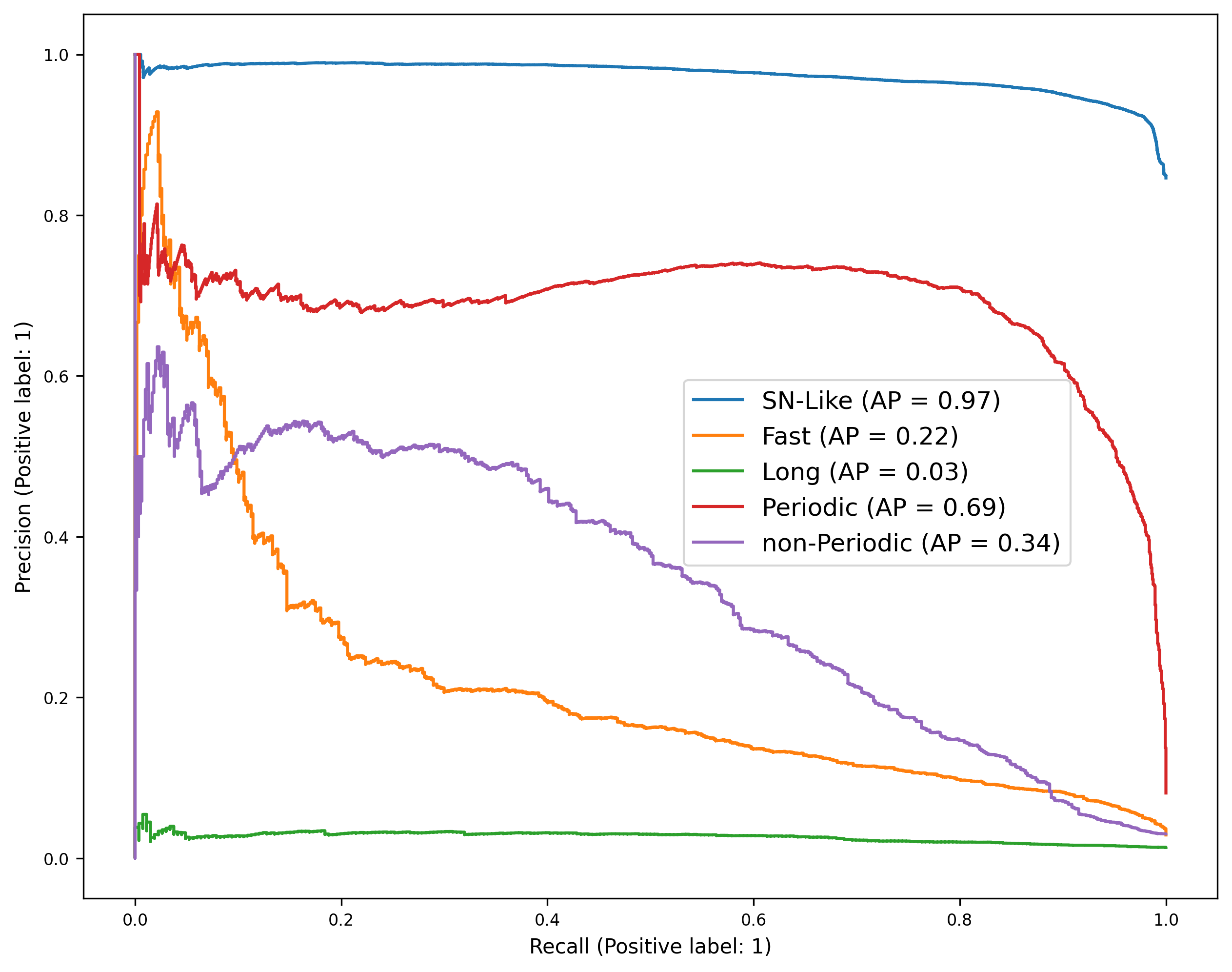}}
    \hfill
    \subfigure[ROC curve for the dataset with a temporal advance of 10 days.]{\includegraphics[width=0.5\textwidth]{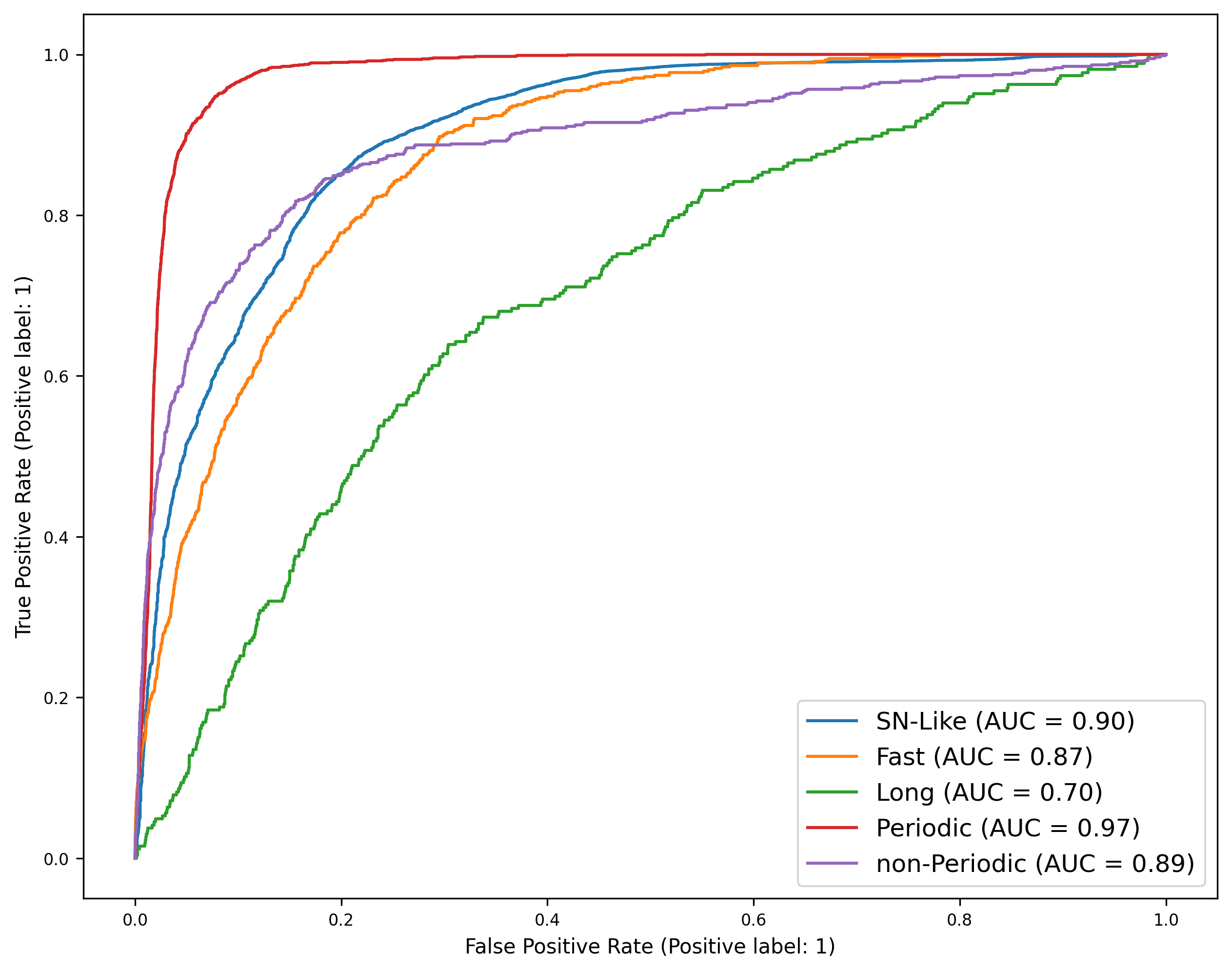}}
    \caption{}
    \label{fig:roc e precision_10}
\end{figure}

\begin{figure}[!h]
    \centering
    \subfigure[Precision-Recall curve for the dataset with a temporal advance of 20 days.]{\includegraphics[width=0.5\textwidth]{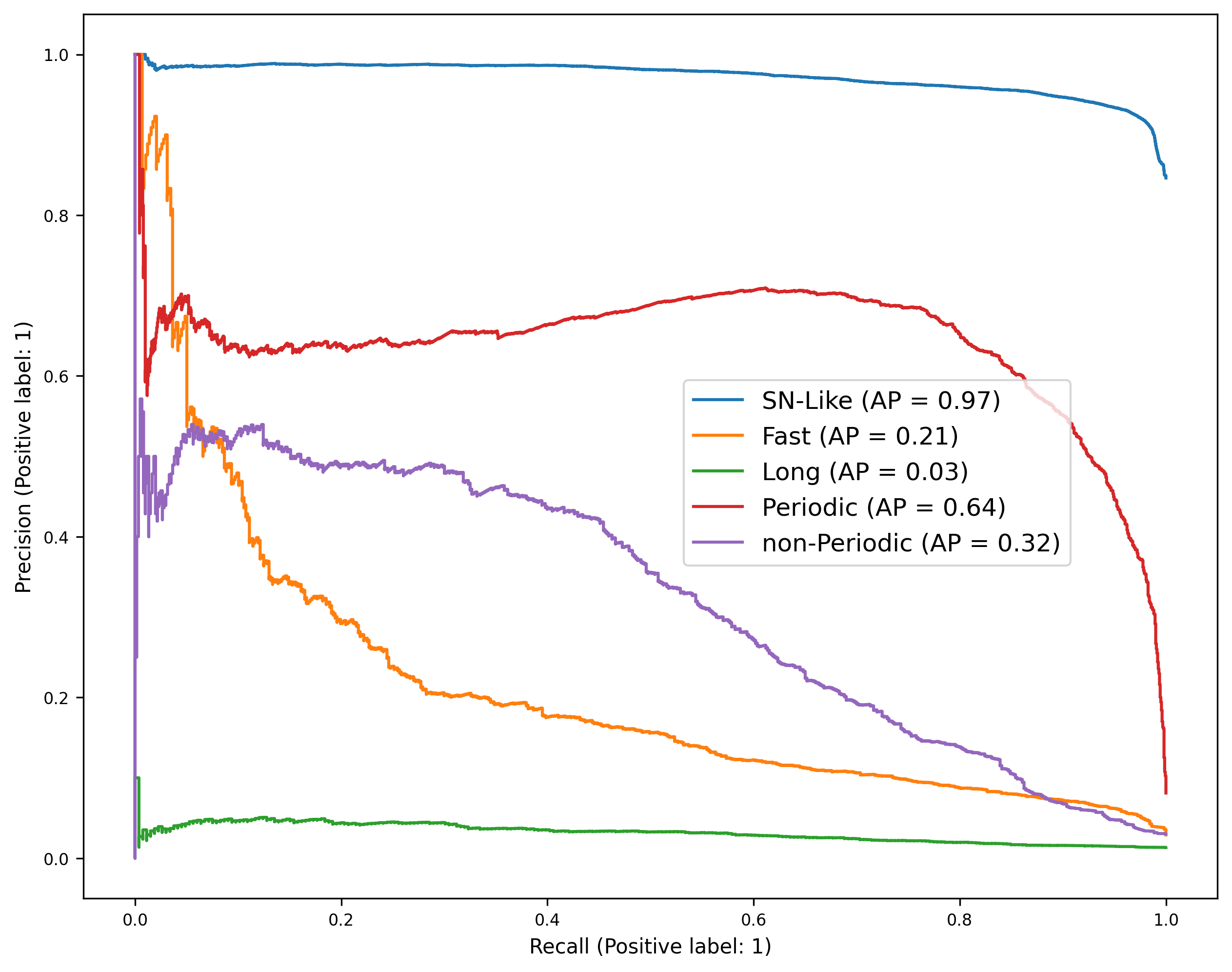}}
    \hfill
    \subfigure[ROC curve for the dataset with a temporal advance of 20 days.]{\includegraphics[width=0.5\textwidth]{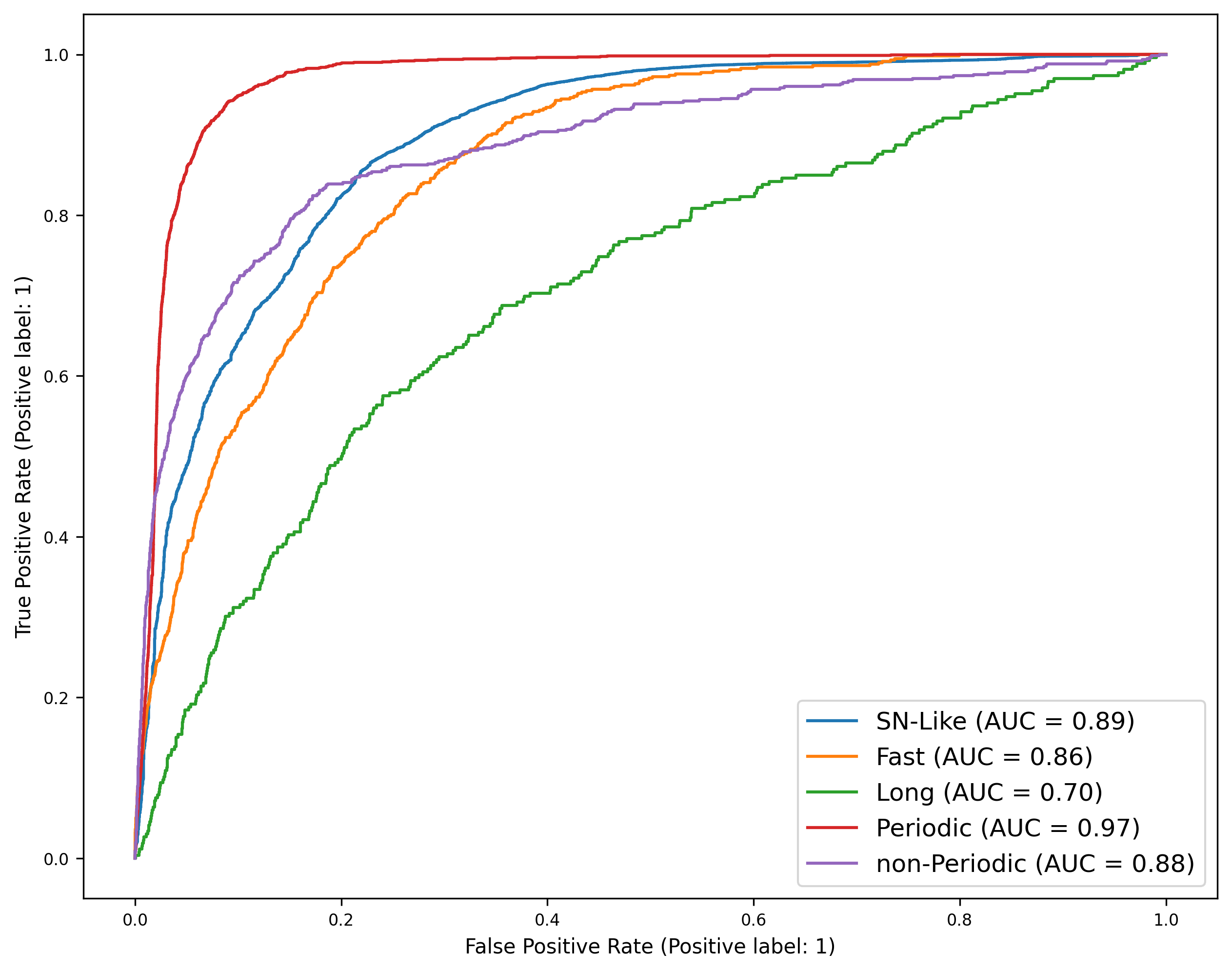}}
    \caption{}
    \label{fig:roc e precision_20}
\end{figure}

If it is of interest to the reader, the ROC and Precision-Recall curves are shown in figures \ref{fig:roc e precision_5}, \ref{fig:roc e precision_10}, and \ref{fig:roc e precision_20} for the test sets of 5, 10, and 20 days.

\end{document}